\documentclass[10pt,twocolumn,letterpaper]{article}

\usepackage{cvpr}      

\usepackage[dvipsnames]{xcolor}

\usepackage{bbding}
\usepackage{graphicx}
\usepackage{graphicx}
\usepackage{amsmath}
\usepackage{amssymb}
\usepackage{booktabs}
\usepackage{bbm}
\usepackage{colortbl}
\usepackage{flushend}
\usepackage[pagebackref=true,breaklinks=true,letterpaper=true,colorlinks,bookmarks=false]{hyperref}
\usepackage[capitalize]{cleveref}
\usepackage{bm}
\usepackage{multirow}
\usepackage{makecell}
\usepackage{tikz}
\usepackage{comment}
\usepackage{amsmath,amssymb} 
\usepackage{color}
\usepackage{colortbl}
\usepackage{tikz}
\usepackage{caption}
\usepackage{hyperref}
\usepackage{url}
\usepackage{times}
\usepackage{epsfig}
\usepackage{graphicx}
\usepackage{amsmath}
\usepackage{amssymb}
\usepackage{bbm}
\usepackage{multirow}
\usepackage{wrapfig}

\ExplSyntaxOn
\newcommand\latinabbrev[1]{
	\peek_meaning:NTF . {
		#1\@}%
	{ \peek_catcode:NTF a {
			#1.\@ }%
		{#1.\@}}} 
\ExplSyntaxOff

\definecolor{verylightgray}{RGB}{234, 250, 254}
\definecolor{fgreen}{RGB}{15, 159, 94}

\definecolor{cvprblue}{rgb}{0.21,0.49,0.74}

\def\paperID{3830} 
\def\confName{CVPR}
\def\confYear{2024}

\title{QDFormer: Towards Robust Audiovisual Segmentation in Complex Environments with Quantization-based Semantic Decomposition}
\author{
Xiang Li$^1$\thanks{This work was done when Xiang Li and Xiaohao Xu were interns at Microsoft.}, Jinglu Wang$^2$, Xiaohao Xu$^3$, Xiulian Peng$^2$, Rita Singh$^1$, Yan Lu$^2$, Bhiksha Raj$^{1,4}$ \\
$^1$ CMU, $^2$ Microsoft Research Asia, $^3$ UMich, $^4$ MBZUAI
}

\def\eg{\latinabbrev{e.g}}
\def\etal{\latinabbrev{et al}}

\begin{document}
\maketitle

\begin{abstract}
Audiovisual segmentation (AVS) is a challenging task that aims to segment visual objects in videos according to their associated acoustic cues. 
With multiple sound sources and background disturbances involved, establishing robust correspondences between audio and visual contents poses unique challenges due to (1) complex entanglement across sound sources and (2) frequent changes in the occurrence of distinct sound events. 
Assuming sound events occur independently, the multi-source semantic space can be represented as the Cartesian product of single-source sub-spaces.
We are motivated to decompose the multi-source audio semantics into single-source semantics for more effective interactions with visual content.
We propose a semantic decomposition method based on product quantization, where the multi-source semantics can be decomposed and represented by several disentangled and noise-suppressed single-source semantics. 
Furthermore, we introduce a global-to-local quantization mechanism, which distills knowledge from stable global (clip-level) features into local (frame-level) ones, to handle frequent changes in audio semantics.
Extensive experiments demonstrate that our semantically decomposed audio representation significantly improves AVS performance, \eg, +21.2\% mIoU on the challenging AVS-Semantic benchmark with ResNet50 backbone.
\end{abstract}

\section{Introduction}
Prompt-guided video object segmentation (VOS) \cite{kirillov2023segment,cheng2023segment,li2023towards,li2024paintseg} focuses on the segmentation of objects in videos according to prompts and predict the object classes if required. The use of various modalities of prompts, such as \textit{text} and \textit{audio}, has promising implications for multi-modal understanding and user-interactive video applications.

\begin{figure}[t]
\centering
\includegraphics[width=\linewidth]{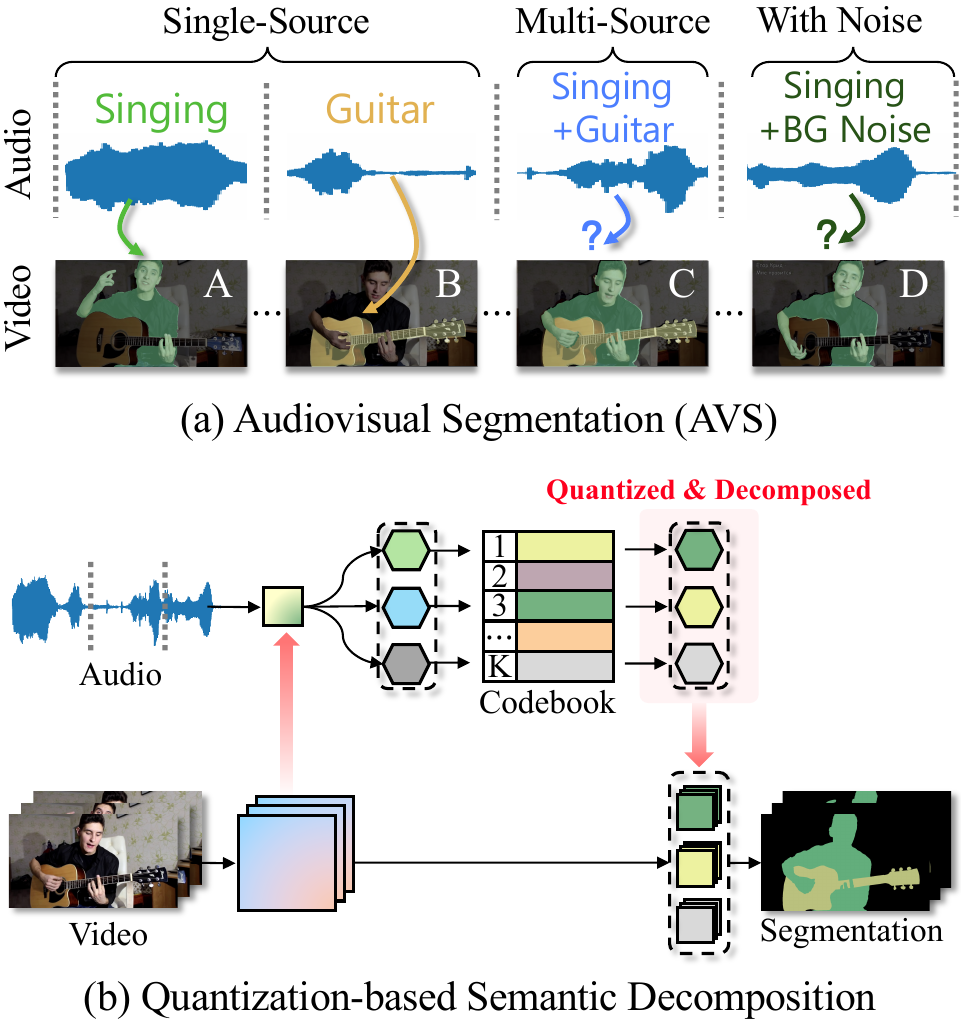}
\vspace{-0.5cm}
\caption{
(a) AVS aims to segment masks associated with sounding objects at each timestamp and predict their classes if required. It presents substantial challenges with multi-source audio (Frame C) or background disturbances (Frame D), due to feature entanglement and noise. (Provided class labels are for visualization only and not included in the input.)
(b)
Our method tackles challenges by decomposing audio features into several semantic tokens. We employ product quantization (with a shared codebook for each decomposed feature) as an information bottleneck to achieve a compact and informative representation.
}
\label{fig:teaser2}
\end{figure}

Numerous recent referring VOS methods \cite{wu2022language,wu2023onlinerefer,referformer,botach2022end} have demonstrated their effectiveness in using linguistic prompts that describe objects discriminatively at a video clip level.
However, employing audio prompts for active-sounding objects in audiovisual segmentation (AVS) \cite{zhou2023audio,zhou2022avs} introduces unique challenges.
First, multiple audio sources could \textit{\textbf{entangle at any timestamp}}, meaning that audio in a single frame could correspond to multiple segmentation masks of different classes, as shown in Frame C of \cref{fig:teaser2} (a). This creates ambiguity in aligning mixed-source audio with visual content, adding complexity to the AVS problem.
Second, audio is often accompanied by \textit{\textbf{background disturbances}}, such as noise or sounds from objects outside the frame, as shown in Frame D of \cref{fig:teaser2} (a).
These disturbances can intensify the difficulties in AVS since acoustic cues are inherently susceptible to such interference. 
As such, directly interacting visual content with audio in complex environments (noisy and entangled) \cite{zhou2022avs,li2023catr,gao2023avsegformer} by imitating clean and discriminative text prompts like \cite{referformer,botach2022end} is not effective. Moreover, adopting pre-trained models for audio separation and noise suppression in advance requires additional data, lacks the flexibility for fine-tuning on AVS-specific datasets, and fails to consider the impact of related visual content.
Addressing these issues by seeking a compact and source-disentangled representation for audio could greatly improve audiovisual interaction and further simplify the AVS problem. 
Also, we believe that exploring solutions to the challenges posed by multi-source and noisy prompts could contribute to the broader field of multi-modal understanding.

\begin{figure}[t]
\centering    
\includegraphics[width=\linewidth]{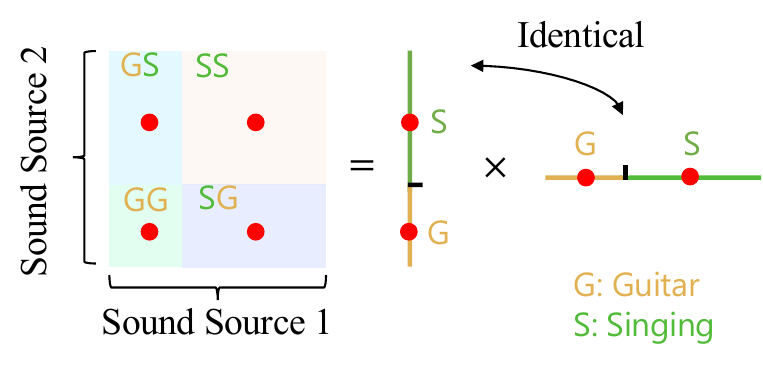}
\vspace{-0.7cm}
    \caption{
    \textbf{An example of decomposing a two-source semantic space.} A multi-source audio semantic space can be assumed as a Cartesian product of two identical single-source semantic spaces, which can be decomposed via product quantization. The {\color{red}{red}} points represent the quantized semantics.
    }
\label{fig:teaser_decompose}
\end{figure}

In this paper, we propose a novel AVS method that leverages the information bottleneck (IB) principle \cite{tishby2000information,alemi2016deep,achille2018emergence} to learn a compact and semantically decomposed audio representation, and then query them to visual features to produce masks, as illustrated in \cref{fig:teaser2} (b).
The IB principle is beneficial as it seeks a compact and task-relevant representation that discards unnecessary details, aids in disentangling underlying factors, and simplifies matching and comparison by converting complex features into discrete symbols.
Specifically, we employ a Product Quantization (PQ) based method to decompose multi-source semantics thanks to its ability to represent a complex space through the product of several subspaces \cite{jegou2010product}.
We define the \textit{audio semantic space} associated with the label set of audio at all timestamps, and the \textit{visual semantic space} associated with the label set of all pixels.
An example of decomposing semantic space is illustrated in \cref{fig:teaser_decompose}.
During single-source moments (Frame A/B of \cref{fig:teaser2} (a)), the visual semantic space corresponds to the semantic labels $\mathcal{Y} = $ \{Guitar (G), Singing (S)\}, and the audio semantic space are of the same label set $\mathcal{Y}$. 
During the two-source moment (Frame C of \cref{fig:teaser2} (a)), the visual semantic space remains consistent to $\mathcal{Y}$, but the \textit{possible} two-source audio semantic space expands to the label set $\mathcal{Y} \times \mathcal{Y} = $\{GG, GS, SG, SS\}. We simplify the AVS problem by assuming independent sound events, which allows us to represent the multi-source semantic space as a Cartesian product of identical single-source semantic spaces. 
By applying product quantization as an information bottleneck and constraining representation dimension to be much smaller than the size of semantic space, achieved by utilizing a shared codebook, we can decompose the multi-source audio semantics into single-source semantics with noise suppression.

Furthermore, it is necessary to analyze audio semantics for every frame, given that sounding sources can change frequently (as the example in \cref{fig:teaser2} (a)). This presents a robustness challenge, as extracted semantics from short-term audio can be unstable compared to the extraction from long-term (clip-level) audio. To improve the frame-level audio representation, we propose a global-to-local mechanism, which distills knowledge from robust global (clip-level) audio representations into local (frame-level) ones. 
Specifically, the codebook is learned with clip-level visual-enriched audio features and performs local quantization on each frame without updating it. 
Thereby, the local semantic tokens are calibrated to the more robust and representative clip-level feature.

In summary, our contribution is three-fold: 
\begin{itemize}
    \item An effective approach of audio semantic decomposition via product quantization, addressing the challenge of correlating audiovisual features introduced by multiple sounding sources and background disturbances.     
    \item A global-to-local distilling mechanism for frame-level audio semantic enhancement, addressing the ineffectiveness of frame-level audio feature extraction.
    \item Extensive experiments are conducted to verify the effectiveness and robustness of the proposed method, which significantly outperforms previous state-of-the-art methods on three AVS benchmarks (+21.2\% mIoU for AVS-Semantic with ResNet-50 backbone).
\end{itemize}
\section{Related Work}
\noindent\textbf{Audiovisual segmentation and localization.}
Audiovisual segmentation (AVS), which was recently introduced \citep{zhou2022avs}, aims to segment the objects that produce sound at the time of the image frame. Zhou \textit{et al.} \citep{zhou2022avs} proposed a method with cross-modal attention to locate the sound source, making it the pioneering work in AVS. Recently, an extended task of AVS, audiovisual semantic segmentation (AVSS), is proposed by Zhou \textit{et al.} \citep{zhou2023audio} which aims to not only segment the mask of sound sources but also predict the category of each sound source. Due to the semantic entanglement in audio, tackling multi-source AVSS is more challenging than AVS task. Zhou \textit{et al.} \citep{zhou2023audio} follows the TPAVI module in \citep{zhou2022avs} to conduct audiovisual interaction. More recently, CATR \cite{li2023catr} introduces an encoding-decoding framework CATR that presents a novel spatial-temporal audio-video fusion block to fully consider the audio-visual combinatorial dependence in a decoupled and memory-efficient manner. Chen \textit{et al.} \cite{yuanhong2023closer} creates a novel benchmark called VPO which augments images from COCO \cite{lin2014coco} with VGGSound \cite{chen2020vggsound} to facilitate the AVS research. 
Sound source localization (SSL) \citep{mo2022closer,mo2022localizing,senocak2018learning,hu2019deep,qian2020multiple,chen2021localizing,afouras2020self} is a related problem to AVS that aims to locate the regions of sounds in the visual frame. Common SSL methods \citep{arandjelovic2018objects,arandjelovic2017look,cheng2020look,senocak2018learning} leverage cross-modal correspondence between audio and visual features to locate sounds, which are then displayed as heatmaps. For instance, Mo \textit{et al.} \citep{mo2022closer} leverage multi-level audiovisual contrastive learning \cite{zhang2023patch1,zhang2022contextual,zhang2023patch,zhang2022align,zhang2021zero,zhang2024continuous} to effectively locate the objects. Different from previous methods primarily designed for single-source scenarios, our objective is to address the semantic entanglement present in multi-source audios and explore methods for effective interaction between multi-source audios and videos.

\vspace{0.2cm}\noindent\textbf{Video Segmentation.}
Audiovisual segmentation is intricately connected to video object segmentation (VOS) \cite{yang2018efficient,jain2017fusionseg,cheng2021modular,seong2020kernelized,hu2021learning,cheng2021stcn,seong2021hierarchical,yang2021associating,li2022hybrid,li2023panoramic,li2022video,zhao2022semantic,zhao2022alignment} and video semantic segmentation (VSS) \cite{li2018low,sun2022coarse,zhuang2022semi,hu2020temporally,paul2020efficient}. Notably, a most closely related task in the video segmentation domain is referring video object segmentation (R-VOS) \cite{botach2021mttr,referformer,urvos,wu2022multi,li2022you,wu2023onlinerefer,tang2023temporal,miao2023spectrum,chen2022multi,li2023robust,han2023html}, which seeks to segment objects in visual frames based on linguistic expressions. In R-VOS, each expression typically refers to a single object, akin to the single-source AVS setting. Recently, \citet{ding2023mevis} introduced the MeViS dataset, expanding the R-VOS task to a more generalized setting that allows referring to multiple objects through a single language expression.


\section{Method}

\begin{figure*}[t!]
    \centering    
    \vspace{-0.1cm}
    \includegraphics[width=\linewidth]{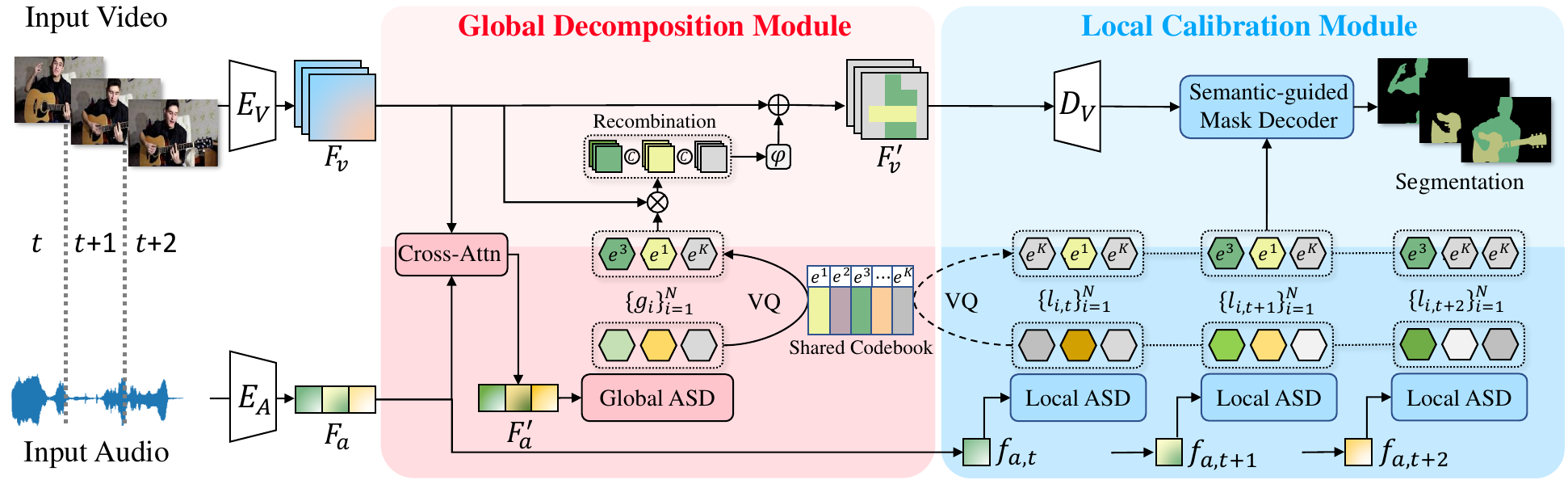}
    \caption{\textbf{Method overview.} Our pipeline accepts an input video along with its associated and non-separated audio, proceeds to segment masks for each frame based on the active audio, and predicts semantic labels of masks if required. 
    The \textbf{global decomposition} module decomposes multi-source audio features $F'_a$ and quantizes all decomposed single-source audio features $\{g_i\}$ with a shared codebook. Visual features $F_v$ are further fused with decomposed audio features to be $F'_v$.
    The \textbf{local calibration} module distills knowledge from stable global (clip-level) audio features $\{\mathrm{VQ}(g_i)\}$ to local (frame-level) audio features $\{\mathrm{VQ}(l_{t,i})\}$ by utilizing the shared codebook. Masks for each frame are generated by the semantic-guided mask decoder with local audio features $\{\mathrm{VQ}(l_{t,i})\}$ and the fused visual feature $F'_v$.
    }
    \label{fig:pipeline}
\end{figure*}

In this section, we first present the formulation of the quantization-based method for multi-source audio semantic decomposition. Then, we outline the pipeline that utilizes the quantized and decomposed audio representation to improve the audiovisual segmentation tasks.

\subsection{Quantization-based Semantic Decomposition}

Without loss of generality, let us consider the $N$-source audio case at a specific timestamp, where each source corresponds to a label in the single-source label set $\mathcal{Y}_s$. 
Due to the entanglement, the multi-source audio feature $x \in \mathcal{X}_m$ will correspond to a much larger label set $\mathcal{Y}_m$, where $\mathcal{X}_m$ denotes the multi-source audio feature space. 

Intuitively, $x$ containing information from $N$-source audio is highly entangled and not effective for interacting with visual content.
Our objective is to seek a decomposition transformation $\pi(x) = (x_1, \cdots, x_N)$. 
We expect that the decomposed $N$ representations $(x_1, \cdots, x_N)$ can be disentangled, such that each $x_i$ can retain the most relevant information to the single-source label in $\mathcal{Y}_s$ while suppressing extraneous and entangled noise.
Inspired by the information bottleneck principle, which facilitates data compression to encapsulate the most task-relevant information and break down the data into independent factors, we propose to apply a quantization-based semantic decomposition method. 

Assuming the independence of sound events, $\mathcal{Y}_m$ becomes a Cartesian product of several equivalent single-source label sets $\mathcal{Y}_s$. We have:
\begin{equation}
\mathcal{Y}_m=\underbrace{\mathcal{Y}_s \times \cdot\cdot\cdot \times \mathcal{Y}_s}_N.
\end{equation} 
Knowing the desired $x_i$ should effectively represent the semantics within $\mathcal{Y}_s$, we construct an information bottleneck by constraining each $x_i$ for $ i=1,\cdots,N$, to exist within the same space, a space with cardinality $K$ roughly equivalent to the size of $\mathcal{Y}_s$. 
Specifically, we perform the product quantization~\cite{jegou2010product} on $x\in\mathcal{X}_m$ with $N$ shared vector quantizer $\mathrm{VQ}(\cdot)$ as:
\begin{equation}
\scalebox{1}{
    $\mathrm{PQ}(x)=\mathrm{VQ}(x_1)\oplus\cdots\oplus\mathrm{VQ}(x_N),  \textrm{ \space\space } \mathrm{VQ} \sim \mathcal{C}$,
}
\end{equation}
where $\oplus$ denotes channel-wise concatenation for re-combining the quantized features, $\sim$ denotes that the vector quantizer $\mathrm{VQ}(\cdot)$ is associated with the codebook $\mathcal{C}=\{e^k\}_{k=1}^K$, that is, $\mathrm{VQ}(\cdot)$ maps a feature $x_i$ to a codeword $e_i=\arg\min_{e^k \in \mathcal{C}}\|x_i-e^k\|_p$ that minimizes the distance between $x_i$ and $e^k \in \mathcal{C}$ in the $p$-norm sense. In this way, with task supervision to enforce $\mathrm{PQ}(x)$ to represent the same information as $x$, we can effectively obtain decomposed $x_i$ using the constructed information bottleneck.

\begin{figure}[t]
    \centering
    \includegraphics[width=\linewidth]{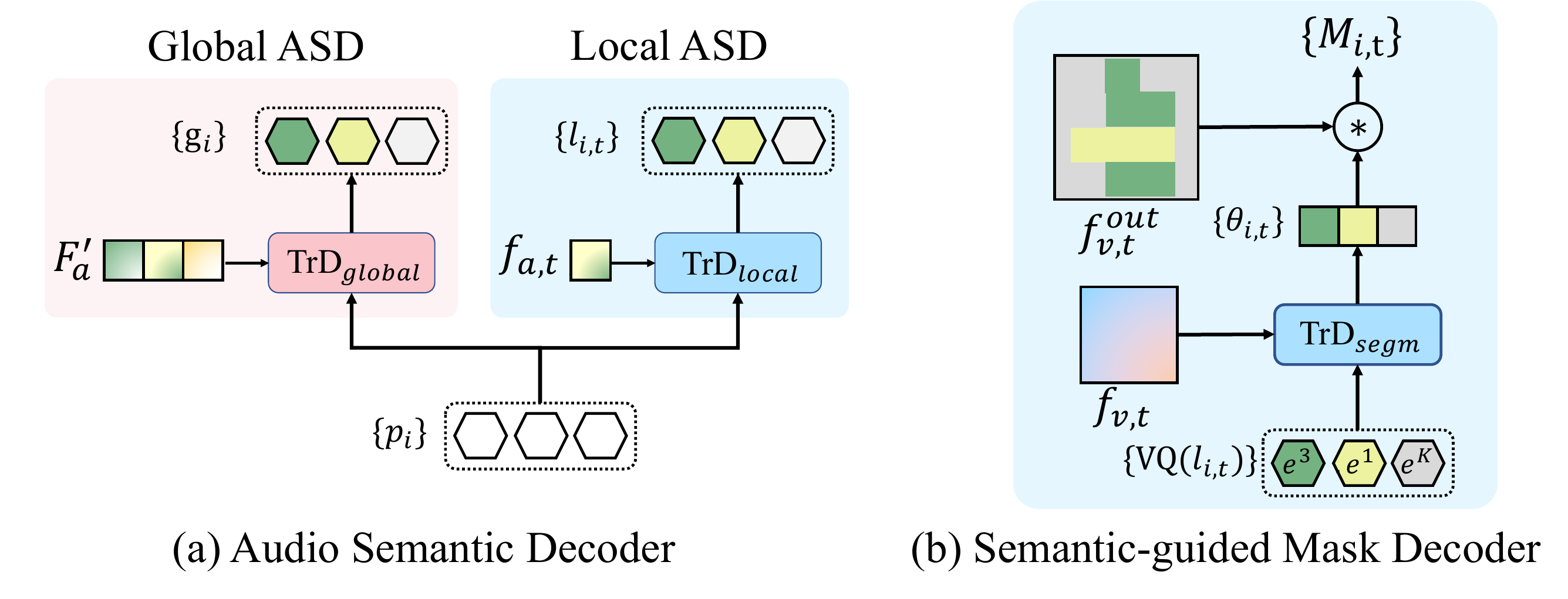}
    \vspace{-0.5cm}
    \caption{
    (a) Global and local audio semantic decoder (ASD) share similar structures that query clip-/frame-level audio features, $F_a^\prime$ or $f_{a,t}$, with a transformer decoder $\mathrm{TrD}_{global}/\mathrm{TrD}_{local}$ using learnable semantic prototypes $\{p_i\}$. (b) The semantic-guided mask decoder contains a transformer decoder $\mathrm{TrD}_{segm}$ to align audiovisual features and computes dynamic filters $\theta_{i,t}$.
    The final mask $M_{i,t}$ is generated by a dynamic convolution between the visual feature $f_{v,t}^{out}$ and $\theta_{i,t}$.
    }
    \label{fig:decoder}
\end{figure}

\subsection{Network Overview}
The proposed framework consists of three main components as illustrated in \cref{fig:pipeline}: feature encoding, global decomposition, and local calibration. 

First, we extract visual features $F_v=\{f_{v,t}\}_{t=1}^T$ with a visual encoder (a backbone with a transformer encoder on top of the backbone), and acoustic features $F_a=\{f_{a,t}\}_{t=1}^T$ with VGGish \cite{hershey2017cnn} (the same setting as previous AVS methods \citep{zhou2022avs, li2023catr}). 
Then, {to decompose semantics in multi-source audio features}, we use a global semantic decomposition module to map the audio query into a set of semantic tokens $\{g_i\}_{i=1}^{N}$. We learn a semantic codebook to quantize them. The quantized tokens are further employed to modulate the visual features to inject information about corresponding sound sources.
Afterward, {to obtain frame-level audio features to query object masks}, we utilize a local semantic decomposition module for each timestamp, which uses the global codebook to decouple local audio semantics. Each quantized local semantic token ${\mathrm{VQ}(l_{i,t})}$ serves as a query to segment a frame-level mask with the semantic-guided mask decoder.
 
\subsection{Global Decomposition}
To tackle the mixture of multi-source audio queries and effectively conduct audiovisual fusion, we propose the global decomposition module to decompose audio semantics at the video clip level, illustrated in the pink box of \cref{fig:pipeline}. It consists of two stages: global semantic decomposition (shown in the lower part) and audiovisual semantic recombination (shown in the upper part).

\noindent\textbf{Global semantic decomposition.}
Global semantic decomposition aims to decompose multi-source audio semantics into single-source semantics. The audio feature $F_a$ is first fused with video feature $F_v$ to be $F'_a$, taking the form:
\begin{equation}
\centering
\begin{aligned}
    F_a^\prime&=\mathrm{LN}(\mathrm{FFN}(h_a)+h_a),\\ 
    h_a=&\mathrm{LN}(\mathrm{MCA}(F_a,F_v)+F_a),
\end{aligned}
\end{equation}
where $\mathrm{MCA}$ denotes Multi-head Cross-Attention, $\mathrm{LN}$ denotes Layer Normalization, and $\mathrm{FFN}$ denotes Feed-Fordward Network. 
After that, we transform the audio feature $F_a^\prime$ to $N$ decomposed semantic tokens $\{g_i\}_{i=1}^N$ with a global audio semantic decoder (global ASD),
\begin{equation}
    g_i=\mathrm{TrD}_{global}(p_i|F_a^\prime),
\end{equation}
by querying a set of learnable semantic prototypes $\{p_i\}_{i=1}^N$ to the feature $F^\prime_a$, with a transformer decoder $\mathrm{TrD}_{global}$ as shown in \cref{fig:decoder} (a).
Each semantic token is then quantized to be $e_i = \mathrm{VQ}(g_i)$ with the shared codebook $\mathcal{C} = \{e^k\}_{k=1}^{K} $, imposing that all semantic tokens to share an identical feature subspace with low cardinality.
Note that we set the codebook size $K \approx |\mathcal{Y}|$ to force the network to learn decomposed semantics, where $|\mathcal{Y}|$ denotes the number of sound event categories. 

\noindent\textbf{Audiovisual semantic recombination.}
Audiovisual semantic recombination aims to leverage the decomposed audio feature to interact with visual features. After obtaining quantized global semantic tokens $\{e_i\}_{i=1}^N$, which encode $N$ groups of decomposed semantics, we aim to interact them with visual features while preserving the original function of the multi-source audio input. A set of dynamic filters $\{w_i\in\mathbb{R}^{D_v}\}_{i=1}^N$ are first learned from global semantic tokens $\{g_i\}_{i=1}^N$ by two linear layers. $D_v$ is the channel dimension of $F_v$. After that, we utilize channel-wise attention to modulate video features by each filter to interact the visual feature with the content referred by different semantic tokens, which is given by:
\begin{equation}
    F_v^\prime=\mathrm{BN}(\varphi(w_i F_v \oplus\cdots\oplus w_N F_v )+F_v ),
\end{equation}
where $\varphi$ denotes a convolution layer to reduce channel from $N\times D_v$ to $D_v$, $\mathrm{BN}$ denotes Batch Normalization, and $\oplus$ denotes concatenation among channels. 
By incorporating channel-wise attention, the visual features can be more effectively concentrated on the relevant audio content. Furthermore, through channel-wise concatenation, the decomposed audio semantics can be reintegrated, producing hybrid semantics that refer to the holistic contents of the original audio input.

\begin{table*}[t]
\centering
\scalebox{0.95}{
\begin{tabular}{l|p{2.2cm}<{\centering}|p{1cm}<{\centering}|p{1cm}<{\centering}p{1cm}<{\centering}|p{1cm}<{\centering}|p{1cm}<{\centering}p{1cm}<{\centering}|p{2.2cm}<{\centering}p{1cm}<{\centering}} 
\hline
\multirow{2}*{Method} & \multirow{2}*{Backbone} & \multicolumn{3}{c|}{AVS-Object-Single} &
\multicolumn{3}{c|}{AVS-Object-Multi} & AVS-Sementic\\
\cline{3-4}\cline{5-9}
~ &&$\mathcal{J} \& \mathcal{F}\uparrow$ & $\mathcal{J}\uparrow$ & $\mathcal{F}\uparrow$ & $\mathcal{J} \& \mathcal{F}\uparrow$ & $\mathcal{J}\uparrow$ & $\mathcal{F}\uparrow$ & mIoU$\uparrow$\\
\hline
\multicolumn{9}{c}{ResNet Backbone} \\
\hline
LVS \citep{chen2021localizing} & ResNet-18 & 44.9 & 37.9 & 51.9 & 31.3 & 29.5 & 33.0 & -\\
MSSL \citep{qian2020multiple} & ResNet-18 & 55.6 & 44.9 & 66.3 & 31.2 & 26.1 & 36.3 & -\\
3DC \citep{mahadevan2020making} & 3DC & 66.5 & 57.1 & 75.9 & 43.6 & 36.9 & 50.3 & 17.3\\
AOT \citep{yang2021associating} & ResNet-50 & - & - & - & - & - & - & 25.4 \\
AVS \citep{zhou2023audio} & ResNet-50 & 78.8 & 72.8 & 84.8 & 52.9 & 47.9 & 57.8 & 20.2\\
Bi-Gen \citep{hao2023improving} & ResNet-50 & 79.8 & 74.1 & 85.4 & 53.4 & 50.0 & 56.8 & - \\
AVSegFormer \citep{gao2023avsegformer} & ResNet-50 & 81.2 & 76.5 & 85.9 & 56.2 & 49.5 & 62.8 & 24.9\\ 
CATR \citep{li2023catr} & ResNet-50 & 81.0 & 74.9 & \bf87.1 & 59.4 & 53.1 & \bf65.6 & -\\
\rowcolor{verylightgray}\textbf{QDFormer} & ResNet-50 & \bf81.8 & \bf77.6 & 86.0 & \bf61.6 & \bf59.6 & 63.5 & \bf46.6\\
\hline
\multicolumn{9}{c}{Transformer Backbone} \\
\hline
iGAN \citep{mao2021transformer} & Swin-Tiny & 69.7 & 61.6 & 77.8 & 48.7 & 42.9 & 54.4 & -\\
SST \citep{duke2021sstvos} & SSL & 73.2 & 66.3 & 80.1 & 49.9 & 42.6 & 57.2 & -\\
LGVT \citep{zhang2021learning} & Swin-Tiny & 81.1 & 74.9 & 87.3 & 50.0 & 40.7 & 59.3 & -\\
AVS \citep{zhou2023audio} & PVT-v2-Base & 83.3 & 78.7 & 87.9 & 59.3 & 54.0 & 64.5 & 29.8\\ 
\rowcolor{verylightgray}\bf QDFormer & Swin-Tiny & \bf83.9 & \bf79.5 & \bf88.2 & \bf64.0 & \bf61.9 & \bf66.1 & \bf53.4\\
\hline
\end{tabular}
}
\vspace{-0.2cm}
\caption{\textbf{Quantitative comparison to AVS and AVSS methods.} 
$\uparrow$ indicates the larger the better. After consulting authors of AVS \cite{zhou2023audio}, we reached an agreement to adopt mIoU as a more standardized metric for evaluating AVSS since several previous methods incorrectly compared boundary metric $\mathcal{F}$ with mask metric F-score defined in \cite{zhou2023audio}.
}
\label{tab:avs}
\end{table*}

\subsection{Local Calibration}
Since the audio query is time-variant, global semantic tokens cannot be accurately aligned with visual features at the frame level. To segment audio-queried contents in each frame, we propose the local calibration module (illustrated in the blue box of \cref{fig:pipeline}), consisting of a local semantic decomposition stage (lower part) and a semantic-guided mask decoding stage (upper part).

\noindent\textbf{Local semantic decomposition.}
This stage aims to decompose the semantics encoded in each audio frame.
Similar to the global ASD, the local audio semantic decoder (Local ASD) decodes frame-level semantics with a transformer decoder $\mathrm{TrD_{local}}$ and a set of semantic prototypes $\{p_i\}_{i=1}^N$, as shown in \cref{fig:decoder} (a). The local semantic tokens $l_{i,t}$ are given by 
\begin{equation}
l_{i,t}=\mathrm{TrD}_{local}(p_i|f_{a,t}).
\end{equation}
where $f_{a,t}$ is $t$-th frame of the audio feature $F_a$.
The local semantic tokens do not build their own codebook but utilize the global codebook $\mathcal{C}$, that is, they do not update $\mathcal{C}$ but are committed to being close to the vectors in $\mathcal{C}$. 
In this way, the local semantic tokens can be calibrated according to global ones, which are more reliable.
Further explanation regarding supervision will be provided in \cref{sec:loss}.

\noindent\textbf{Semantic-guided mask decoding.}
We utilize the semantic-guided mask decoder to decode visual features into masks that correspond to decomposed local audio semantics, with the detailed structure illustrated in \cref{fig:decoder} (b).
Pyramid video features $F_v^{out}=\{f_{v,t}^{out}\}_{t=1}^T$ are obtained with the feature pyramid network \citep{lin2017fpn}.
We leverage a shared multimodal transformer decoder $\mathrm{TrD}_{segm}$ to generate dynamic filters $\theta_{i,t}=\phi_{segm}(\mathrm{TrD}_{segm}(\mathrm{VQ}(l_{i,t})|f_{v,t}))$ for each timestep, where $\phi_{segm}$ is a two-layer fully-connected network. The final mask segmentation can be obtained by:
\begin{equation}
    M_{i,t}=f^{out}_t\ast \theta_{i,t},
\end{equation}
where $\ast$ denotes the dynamic convolution \citep{chen2020dynamic}. Each filter represents semantics of a decomposed single-source audio, contributing to the segmentation of the single sounding object. Additional class probability prediction $P_{i,t}$ and bounding box prediction $B_{i,t}$ for each mask $M_{i,t}$ are performed by two two-layer fully connected networks from the output of $\mathrm{TrD}_{segm}(\mathrm{VQ}(l_{i,t})|f_t)$.

\subsection{Loss Function}
\label{sec:loss}
The overall loss function is given by 
\begin{equation}
\mathcal{L}=\lambda_{quant}\mathcal{L}_{quant}+\mathcal{L}_{segm},
\end{equation}
where $\mathcal{L}_{quant}$ and $\mathcal{L}_{segm}$ are the loss for semantic quantization and segmentation, respectively. $\lambda_{quant}$ is a constant.

\noindent\textbf{Loss for semantic quantization.}
The quantizer is shared with both global and local semantic decomposition, while the local semantic tokens do not update the codebook. The loss is given by
\begin{equation}
{\fontsize{8.5pt}{\baselineskip}\selectfont
\begin{aligned}
    \mathcal{L}_{quant} = &\sum_{i=1}^N\Big\{\|\mathrm{VQ}(g_i)-\mathrm{sg}[g_i]\|_2^2 + \lambda_{com}\|\mathrm{sg}[\mathrm{VQ}(g_i)]-g_i\|_2^2\\ + &\lambda_{com}\sum_{t=1}^T\|\mathrm{sg}[\mathrm{VQ}(l_{i,t})]-l_{i,t}\|_2^2\Big\},
\end{aligned}}
\end{equation}
where $\mathrm{sg} \left [ \cdot \right ]$ stands for stop-gradient operation. $\mathrm{VQ}(\cdot)$ denotes the vector quantization function, where $\mathrm{VQ}(x)=e_i = \arg\min_{e^k}\|x-e^k\|_2 \in \mathcal{C}$ and $\mathcal{C}=\{e^k\}_{k=1}^{K}$ is the shared codebook. $\lambda_{com}$ is a constant. The first term aims to update the codebook. The second and third terms aim to minimize the quantization error by forcing the input vector to be quantized to its closest vector in the codebook.

\noindent\textbf{Loss for segmentation.}
Let the predictions of the network be $\mathrm{\mathbf{y}}=\{\mathrm{\mathbf{y}}_{i}\}_{i=1}^{N}$ where $y_i=\{B_{i,t}, P_{i,t}, M_{i,t}\}_{t=1}^{T}$. $B_{i,t}$, $P_{i,t}$ and $M_{i,t}$ denote bounding box, class probability and mask predictions, respectively. We denote the ground-truth as $\hat{\mathrm{\mathbf{y}}}=\{\hat{\mathrm{\mathbf{y}}}_j\}_{j=1}^N$ (padded with $\emptyset$ \citep{cheng2021mask2former}) where $\hat{y}_j=\{\hat{B}_{j,t}, \hat{C}_{j,t}, \hat{M}_{j,t}\}_{t=1}^{T}$. $C_{j,t}$ is the ground-truth class for the $j$-th sounding object in the video at $t$ frame. We search for an assignment $\sigma \in\mathcal{P}_N$ with the highest similarity where $\mathcal{P}_N$ is a set of permutations of N elements. The similarity can be computed as 
{
\fontsize{9.5pt}{\baselineskip}\selectfont
\begin{equation}
    \begin{aligned}
    \mathcal{L}_{match}(y_i,\hat{y}_j) = \lambda_{box}\mathcal{L}_{box}+\lambda_{cls}\mathcal{L}_{cls}+\lambda_{mask}\mathcal{L}_{mask},
    \end{aligned}
\end{equation}
}
where $\lambda_{box}$, $\lambda_{cls}$, and $\lambda_{mask}$ are weights to balance losses. We leverage a combination of Dice \citep{li2019dice} and BCE loss as $\mathcal{L}_{mask}$, focal loss \citep{lin2017focal} as $\mathcal{L}_{cls}$, and GIoU \citep{rezatofighi2019giou} and L1 loss as $\mathcal{L}_{box}$. The best assignment $\hat{\sigma}$ is solved by the Hungarian algorithm \citep{kuhn1955hungarian}. Given the best assignment $\hat{\sigma}$, the segmentation loss between ground-truth and predictions is defined as $\mathcal{L}_{segm}=\mathcal{L}_{match}(y_i, \hat{y}_{\hat{\sigma}(j)})$.

\section{Experiments}

\noindent\textbf{Dataset.}
We conduct experiments on AVS-Object \citep{zhou2022avs} for AVS task and AVS-Semantic \citep{zhou2023audio} for AVSS task.
\begin{itemize}
    \item \textbf{AVS-Object}: AVS-Object dataset contains 5,356 short videos with corresponding audios in which 4,932 audios contain single-source and 424 audios contain multiple sources. Class-agnostic masks are given as annotations for AVS task. Typically, it is evaluated separately for single- and multi-source audios as \textbf{AVS-Object-Single} and \textbf{AVS-Object-Multi}. 
    \item \textbf{AVS-Semantic}: AVS-Semantic is an extended dataset from AVS-Object which contains 12,356 videos with 70 classes. Semantic segmentation is annotated for AVSS task. Both single- and multi-source audio cases exist in the AVS-Semantic.
\end{itemize}

\begin{figure*}[t]
    \centering    \includegraphics[width=0.85\linewidth]{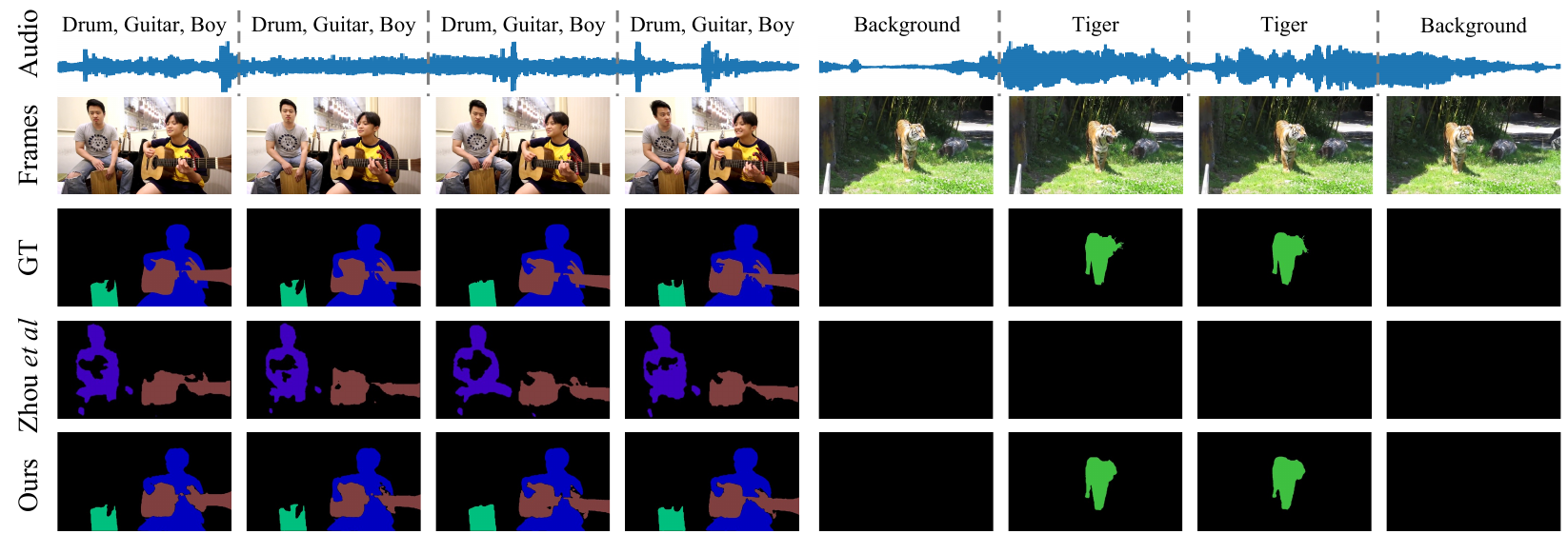}
    \vspace{-0.15cm}
    \caption{Qualitative comparison to Zhou \etal$\,$ \citep{zhou2023audio} on AVS-Semantic. Each color represents a semantic category. Note that we show the class labels in the first row as references but not given in the input, and audio signals and frames are the input.}
    \label{fig:vis}
\end{figure*}

\begin{table*}
\centering
\scalebox{0.95}{
\begin{tabular}{
c|lp{1.4cm}
<{\centering}p{1.4cm}<{\centering}p{1.4cm}<{\centering}|p{3cm}<{\centering}} 
\hline
\multirow{2}*{ID} & \multirow{2}*{Method} & \multicolumn{3}{c}{AVS-Object-Multi} & AVS-Sementic\\
\cline{3-6}
~ & ~ & $\mathcal{J} \& \mathcal{F}\uparrow$ & $\mathcal{J}\uparrow$ & $\mathcal{F}\uparrow$ & mIoU$\uparrow$\\
\hline
1 & Baseline (w/o decomposition)  & 52.9\color{fgreen}{$_{-8.7}$} & 50.1 & 55.7 & 33.5\color{fgreen}{$_{-13.1}$} \\
\hline
2 & + TD \cite{chen2022zero} & 56.2\color{fgreen}{$_{-5.4}$}  & 54.5 & 57.9 & 38.9\color{fgreen}{$_{-7.7}$}  \\
\hline
3 & + Non-Q-based-SD  & 57.6\color{fgreen}{$_{-4.0}$} & 56.1 & 59.1 & 39.4 \color{fgreen}{$_{-6.8}$} \\
4 & + Q-based-SD & 59.7\color{fgreen}{$_{-1.9}$} & 57.6 & 61.8 & 42.5\color{fgreen}{$_{-4.5}$} \\
5 & + Q-based-SD + AVSR   & 60.1\color{fgreen}{$_{-1.5}$}  & 58.2 & 61.9 & 44.5\color{fgreen}{$_{-2.1}$} \\
\rowcolor{verylightgray} 6 & + Q-based-SD + AVSR + LC (\textbf{ours}) & 61.6 & 59.6 & 63.5 & 46.6 \\
\hline
\end{tabular}
}
\label{tab:ablation}
\vspace{-0.05cm}
\caption{\textbf{Component effectiveness study.} TD: decompose audio at the time domain in advance with an off-the-shelf model \cite{chen2022zero}; Non-Q-SD: semantic decomposition without quantization; Q-based-SD: quantization-based semantic decomposition; AVSR: audiovisual semantic recombination; LC: local calibration.}
\label{tab:ablation}
\vspace{-0.4cm}
\end{table*}

\noindent\textbf{Metrics.}
For AVS task, the convention is to compute region similarity $\mathcal{J}$ and contour accuracy $\mathcal{F}$ as defined in \citep{pont2017davis}. Note that we follow the video segmentation convention to use the region similarity $\mathcal{J}$, which is equivalent to $\mathrm{mIoU}$ in the binary AVS setting. For AVSS, we follow the semantic segmentation convention to evaluate the model using mIoU which is defined as the intersection over union averaged among all classes.  

\noindent\textbf{Implementation Detail.}
We implement our method in PyTorch \citep{paszke2019pytorch}. We train our model for 13 epochs and 16 epochs with a learning rate multiplier of 0.1 at the $11^{th}$ and $14^{th}$ epochs for AVS-Object and AVS-Semantic, respectively. We set the initial learning rate to be 1e-4, and a multiplier of 0.5 was applied to the backbone. We adopt $\mathrm{batchsize}=4$ and an AdamW \citep{loshchilov2017adamw} optimizer with weight decay $5\times10^{-4}$. The codebook size $K$ is set to 128. The token number $N$ for decomposing audio is set to 5. If no other specification, all images are resized to have the longest side 224 during evaluation. More details are available in the Appendix.

\subsection{Main Results}
\noindent\textbf{Quantitative comparison on AVS-Object.}
We compare our method with existing AVS methods with CNN and transformer backbones. Without loss of generality, we adopt RestNet-50 and the small Swin-Tiny backbone, and provide results of larger backbones in the Appendix. Note that PVT-v2-Base is much larger than Swin-Tiny.
Our method outperforms the previous state-of-the-art (SOTA) method AVSegFormer \citep{gao2023avsegformer} and CATR \cite{li2023catr} by 5.4 and 2.2 of $\mathcal{J}\&\mathcal{F}$ score on AVS-Object-Multi datasets respectively (with ResNet-50 backbone). The promising $\mathcal{F}$ score achieved by CATR can be attributed to its additional incorporation of atrous spatial pyramid pooling \cite{chen2017deeplab}, a feature that can be seamlessly integrated into our method as well. We notice that the improvement on the multi-source setting is much larger than the single-source setting. This is because single-source audios contain simple and disentangled semantics and can be easily aligned with visual features while, for multi-source audios, the complex semantic space makes the alignment to visual contents much more difficult.

\noindent\textbf{Quantitative comparison on AVS-Semantic.}
Compared to the AVS-Object task, our method demonstrates greater improvement in the AVS-Semantic task. As shown in the \cref{tab:avs}, our method eclipses the previous SOTA AVSS method AOT \citep{yang2021associating} by a remarkable 21.2 mIoU with ResNet-50 backbone. The improvement in the AVSS task can be attributed to several factors. First, the task itself involves the semantic prediction of sound sources. However, due to mixed audio signals, aligning visual content accurately becomes challenging, leading to difficulties in classification. Secondly, the number of sound sources and categories of AVS-Semantic are larger than AVS-Object, which will result in a larger semantic space. When the mixed semantics are not decomposed, the network struggles to handle the numerous mixed semantics effectively. Thirdly, in the AVS-Semantic dataset, sound event changes occur more frequently. As a result, a more robust frame-level audiovisual correspondence is required. Our proposed global-to-local distilling mechanism addresses this challenge by enhancing the capture of local semantic information, enabling accurate object segmentation.

\noindent\textbf{Qualitative comparison.}
As shown in \cref{fig:vis}, we qualitatively compare our method to the method proposed by Zhou \etal$\,$ \citep{zhou2022avs} on AVS-Semantic. Our method achieves better results on both segmenting quality and class prediction accuracy. Since the method \citep{zhou2022avs} directly fuses mixed audio features with video features, we notice that it suffers from object incorrectness when multiple sound sources are present. Meanwhile, due to the lack of frame-level audiovisual calibration, \citep{zhou2022avs} cannot effectively handle the audio semantic changes. More qualitative results are available in the Appendix and supplemental video.

\subsection{Ablation Study}
We carry out ablation studies on multi-source audio scenarios using the AVS-Object-Multi and AVS-Semantic datasets. These studies aim to assess the effectiveness of various components and design choices in our proposed method for quantization-based semantic decomposition, as presented in \cref{tab:ablation}. We use the ResNet-50 backbone for all ablation studies. 

\noindent\textbf{Effectiveness of audio decomposition.}
We establish a baseline model that includes only the two unimodal encoders, cross-attention for fusion, and the mask decoder. This model does not incorporate any decomposition but performs audiovisual interaction directly with cross-attention layers. The model's performance significantly deteriorates in multi-source audio scenarios, yielding scores of 52.9 $\mathcal{J}\&\mathcal{F}$ for the AVS-Object-Multi dataset and 33.5 mIoU for the AVS-Semantic dataset (Row 1). This is an 8.7 $\mathcal{J}\&\mathcal{F}$ and 13.1 mIoU decrease compared to our semantic decomposition method (Row 6).

\noindent\textbf{Effectiveness of decomposition on the semantic domain.}
As a straightforward method for audio decomposition is to separate audio signals in advance with pre-trained models, we provide an alternative method that first applies a widely used sound source separation model \citep{chen2022zero} and then performs audiovisual segmentation for each decomposed audio using our baseline model. This method has gained 3.3 $\mathcal{J}\&\mathcal{F}$ and 5.5 mIoU from the baseline, which also demonstrates that audio decomposition is helpful.
However, it lags behind our method that decomposes audio in the semantic domain, with a decrease of 5.4 $\mathcal{J}\&\mathcal{F}$ and 7.7 mIoU. We attribute this improvement to two factors: 1) the imperfection of sound source separation model, and 2) the conflicts that arise when combining the masks for each source in the time domain without considering visual content during separation. In contrast, our semantic-domain approach does not suffer from these issues and can effectively leverage the information contained in both audio and visual modalities.

\noindent\textbf{Effectiveness of quantization for semantic decomposition.}
To validate the role of quantization, we set up a model that excludes the codebook learning and preserves all other modules for comparison.
Without quantization, the results exhibit 4.0 $\mathcal{J}\&\mathcal{F}$ and 6.8 mIoU decrease (Row 3). The information bottleneck principle enhances our semantic decomposition by reducing irrelevant noise, clustering semantic-relevant features for disentanglement, and improving the matching and comparison of features in the segmentation task.
In addition, for further fusing with visual features, we perform audiovisual semantic recombination (AVSR) to render the decomposed audio tokens aware of visual features. The results of Row 5 show an increase of 0.4 in  $\mathcal{J}\&\mathcal{F}$ and 2.4 in  mIoU (as is shown in Row 4).

\noindent\textbf{Effectiveness of local calibration.}
We perform the local calibration by enforcing local quantization to align with global ones using a shared codebook and not updating it after local LSD. With the local calibration scheme, the audio representation is learned more robustly with long-term information, gaining 1.5 in $\mathcal{J}\&\mathcal{F}$ and 2.1 in mIoU (as is shown in Row 5).

Ablation on codebook size, semantic token number, frame number, and input resolution are in the Appendix.

\begin{figure}[t]
    \centering

    \includegraphics[width=\linewidth]{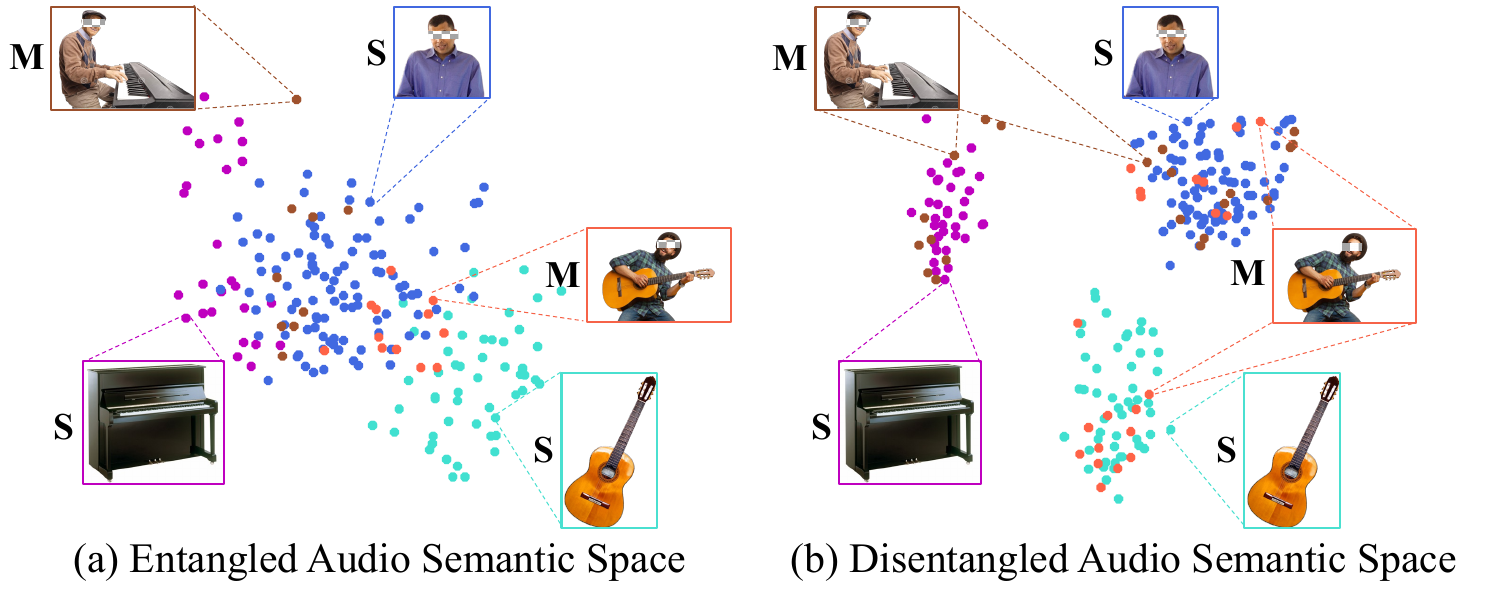}
    \caption{Visualization comparison between entangled and our disentangled audio semantic space. ``M" and ``S" notations denote multi-source and single-source inputs.}
    \label{fig:space}
    \vspace{-0.2cm}
\end{figure}

\subsection{Analysis}
We delve deeper into our research by visualizing the feature space to understand our semantic decomposition better and exploring its robustness against background disturbances.

\noindent\textbf{Visualization of decomposed semantic space.}
As shown in \cref{fig:space}, we visualize the semantic space with and without semantic decomposition on the AVS-Semantic dataset using t-SNE \cite{van2008visualizing}. Three types of single-source audios (``man", ``guitar", ``piano") and two types of multi-source audios (``man+guitar", ``man+piano") are enrolled. Without decomposition, the multi-source features are highly entangled, presenting less evidence related to single-source semantics. However, after performing semantic decomposition, the ``man+guitar" feature presents clear evidence related to its corresponding single-source (``man'' and ``guitar'') semantics. This is reflected in the proximity of the ``man+guitar" feature to the centroids of its corresponding single-source features. The same applies to the ``man+piano" feature.


\begin{figure}[t]
\centering
\vspace{-0.1cm}
\includegraphics[width=\linewidth]{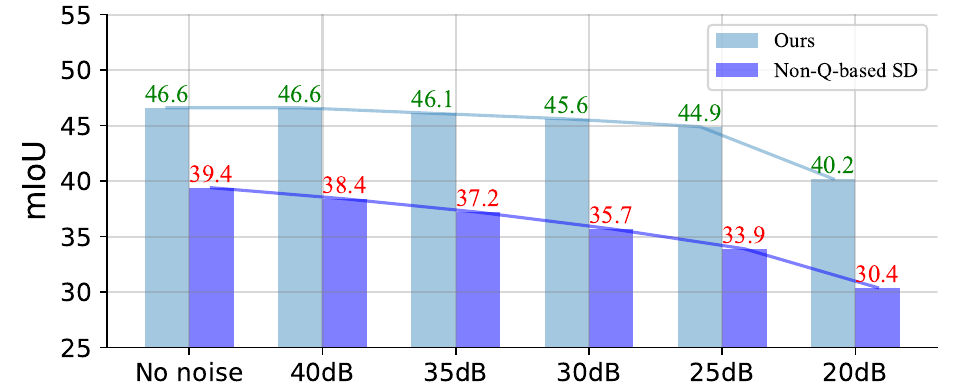}
\vspace{-0.5cm}
    \caption{Analysis with different degrees of background noises (invisible sound sources) on AVSS task. The noise degree is measured by the signal-to-noise ratio (SNR) of the original sound.}
    \label{fig:noise}
    \vspace{-0.2cm}
\end{figure}

\noindent\textbf{Analysis of robustness against background noises.}
We address the susceptibility of audio signals to background disturbances by employing quantized audio representation. In \cref{fig:noise}, we evaluate performance under varying degrees of background noises (drawn from AudioSet \cite{gemmeke2017audio}). As the audio signal-to-noise ratio (SNR) decreases, the model utilizing quantized and decomposed audio representation consistently outperforms the one using conventional continuous representation. This superior performance can be attributed to two factors: (1) effective quantization discards unnecessary noisy details, enhancing audio representation robustness, and (2) PQ-based decomposition disentangles target audio semantics from noisy elements, reducing the impact of background noises in audiovisual interaction. In summary, our approach demonstrates improved resilience to background noise.

\vspace{-0.1cm}
\section{Conclusion}
\vspace{-0.1cm}
This paper explores the quantized and decomposed audio representation to address the audiovisual segmentation. Our approach effectively disentangles multi-source audio semantics with a PQ-based decomposition, providing a more robust audio representation for audiovisual interaction in complex environments. The global-to-local mechanism further enhances frame-level audio semantics, mitigating the instability associated with short-term extractions. Through extensive experiments, our approach has demonstrated superior performance compared to previous methods, particularly excelling in multi-object scenarios. 

\renewcommand{\thetable}{{\Alph{table}}}
\renewcommand{\thefigure}{{\Alph{figure}}}
\renewcommand{\thesection}{{\Alph{section}}}
\setcounter{figure}{0}   
\setcounter{table}{0}   
\setcounter{section}{0}   


{
    \small
    \bibliographystyle{ieeenat_fullname}
    \bibliography{main}
}


\end{document}